%% file: main.tex
\documentclass[letterpaper, preprint, paper,11pt]{AAS}

\usepackage{bm}
\usepackage{amsmath}
\usepackage[colorlinks=true, pdfstartview=FitV, linkcolor=black, citecolor= black, urlcolor= black]{hyperref}
\usepackage{overcite}
\usepackage{footnpag}	

\usepackage{float}
\usepackage{afterpage}
\usepackage{subfig}
\usepackage{placeins}
\usepackage{import}
\usepackage{subfiles}

\graphicspath{{./}{figures/}}
\newcommand{\comment}[1]{}
\begin{document}

\title{FPGA Hardware Acceleration for Feature-Based Relative Navigation Applications}

\author{Ramchander Rao Bhaskara \thanks{Graduate Research Assistant, Aerospace Engineering Department, Texas A\&M University, College Station, TX}, Manoranjan Majji \thanks{Associate Professor, Aerospace Engineering Department, Texas A\&M University, College Station, TX}
}

\maketitle{}

\begin{abstract}

Estimation of rigid transformation between two point clouds is a computationally challenging problem in  vision-based relative navigation. Targeting a real-time navigation solution utilizing point-cloud and image registration algorithms, this paper develops high-performance avionics for power and resource constrained pose estimation framework.   
A Field-Programmable Gate Array (FPGA) based embedded architecture is developed to accelerate estimation of relative pose between the point-clouds, aided by image features that correspond to the individual point sets. At algorithmic level, the pose estimation method is an adaptation of Optimal Linear Attitude and Translation Estimator (OLTAE) for relative attitude and translation estimation. At the architecture level, the proposed embedded solution is a hardware/software co-design that evaluates the OLTAE computations on the bare-metal hardware for high-speed state estimation. The finite precision FPGA evaluation of the OLTAE algorithm is compared with a double-precision evaluation on MATLAB for performance analysis and error quantification. Implementation results of the proposed finite-precision OLTAE accelerator demonstrate the high-performance compute capabilities of the FPGA-based pose estimation while offering relative numerical errors below 7\%. 
\end{abstract}

\section{Introduction} \label{sec:intro}
\input{section1}

\section{Related work}
\input{section2}

\section{Point Cloud Registration}
\input{section3}

\section{Overview of the hardware design}
\input{section4}

\section{Implementation Results}
\input{section5}

\section{Conclusion}
\input{section6}


\bibliographystyle{AAS_publication}   
\bibliography{References, myReference}   

\end{document}

%% file: section1.tex


Emphasis on perception of the 3D world is bringing about a paradigm shift in autonomous navigation and robotics. In avionic applications, 3D imaging through LiDAR and stereo cameras are candidate sensor solutions to realize precision landing \cite{epp2008autonomous, johnson2008overview}, terrain relative navigation during planetary descents \cite{johnson2007general}, proximity operations \cite{jasiobedzki2005autonomous}, robotic servicing or refueling, and hazard avoidance \cite{amzajerdian2016imaging}. These applications are realized through reconstruction of the 3D world, localization in the 3D space or through relative pose estimation \cite{liu2016point}. Inherently, they point to the identification of a rigid transformation between two point-cloud scans, which is termed as a point-cloud registration problem. In a conventional setting, the 3D scanning platforms such as LiDAR gather large amounts of data for registration on a central onboard computer (OBC), which often possess limited computational resources, communication bandwidth, and power budgets. Moreover, in practice, classical iterative solvers for registration, such as Iterative Closest Point (ICP) \cite{besl1992method} methods are not optimized for real-time operation under constrained resources \cite{fortescue2011spacecraft}. The solvers, that minimize the Euclidean distance between individual points, are sensitive to degenerate measurements (outliers, missing, and overlapping points) and result in slow and poor convergence. This motivates a need for (i) improvement in robustness and processing time by fusing 2D image feature data when available \cite{kumar2020lidar}, (ii) a lightweight registration algorithm, and (iii) a low-power low-cost reconfigurable hardware, such as a Field Programmable Gate Array (FPGA), that performs high speed parallel computations for real-time perception.

Geometric methods such as ICP \cite{besl1992method} iteratively minimize the differences between two point-cloud correspondences and in formulation are identical to the Wahba's problem \cite{wahba1965least, arun1987least}. The advent of Optimal Linear Attitude Estimator (OLAE) quickly found its adaptation to the point-cloud registration problem \cite{wong2018optimal}. Along with the attitude, OLAE is also modified to incorporate estimation of the translation vector \cite{majji2010registration, 9206167}. It is this variant of the original OLAE that is referred to as the Optimal Linear Translation and Attitude Estimation (OLTAE) algorithm, which is the main subject of this work. With attitude parameterized using Classical Rodriguez Parameters (CRPs), OLTAE is a rigorously linear system of equations for 6-DOF relative pose estimation. OLTAE's closed-form structure presented in this work brings a better scalability and performance advantage, making it an optimal choice for resource-limited computing platforms. 

This work presents an embedded system design for accelerating the OLTAE algorithm that is potentially implementable on a low-cost FPGA system-on-chip (SoC) platform. Because of the variability involved in the number of measurements and the system output model, the hardware architectures for batch least-squares solvers cannot generally be scalable. As will be shown, OLTAE, at the logical level, is a batch least-squares algorithm that regresses on 2D image feature correspondences across frames, for relative pose estimation. This paper pursues a hardware-scalable approach for pose estimation and presents a modified closed-form solution to the OLTAE algorithm.
\comment{In order to be robust and scale independent, the embedded system implementing the OLTAE logic is equipped to handle a variable number of feature correspondences without considerable design modifications. Since the hardware is difficult to scale, a modified closed-form solution of the OLTAE is presented in this work.}
The modified algorithm uses matrix manipulations to convert the higher dimensional matrix operations involved in a straightforward OLTAE algorithm \cite{majji2010registration, wong2018optimal}, into a problem that gets away with handling only the matrices of dimension three, for any number of incoming measurements. This modified OLTAE not only accelerates the hardware and software performances by operating on less expensive three-dimensional matrix operations, but also consumes lower memory footprint. As will be shown, the formulation in this approach enables handling a variable number of feature measurements without many additional modifications to the hardware architecture, and thereby solves the scaling issue. We use Xilinx Zynq 7020 FPGA SoC to implement the OLTAE logic as a dedicated intellectual property (IP) core \cite{xilix2016zynq1}. Experiments demonstrate that our OLTAE core provides a significant improvement in performance while providing equivalent accuracy in comparison with the mainstream floating-point operations on software. 

The rest of this article is structured as follows: first, a brief survey on FPGA acceleration solutions for point cloud registration is presented. Next, an overview of the point cloud registration process is outlined. In the subsequent section, OLTAE and its modified form for the determination of 6-DOF relative translation and rotation estimation are derived. Then, the hardware architecture for implementing the OLTAE algorithm on an FPGA is presented. Finally, implementation results are presented to qualify the accuracy of our fixed-point hardware design relative to the double-precision software implementation on MATLAB.


%% file: section2.tex

By sheer number of efforts, the development of navigation algorithms so far outweighed their adoption for customized hardware realization due to complexities in hardware design practices as well as search for advancements in optimal algorithms. A need for specific yet scalable hardware designs is of growing importance to add to the library of architectures for easy adoption and application. Investigations on FPGA-based acceleration for 3D relative navigation are seeing an uptick  \cite{ramchander2021hardware,https://doi.org/10.48550/arxiv.2208.03605}.

Sanchez\cite{sanchez2022terrain}, \v{C}{\'\i}{\v{z}}ek et al.\cite{vcivzek2016low}, Nikolic et al.\cite{nikolic2014synchronized}, Konomura et al.\cite{konomura2014visual} discuss low-latency feature detection algorithms for FPGA-based visual odometry. Their respective methods discuss acceleration of image processing techniques for detection of image keypoint features suitable to their respective applications (SIFT, FAST, Harris, and other types). Because feature detection is strongly correlated to the texture of the object (or) terrain that is imaged, our work relies on computer vision libraries\cite{bradski2008learning} to rapidly adapt to different types of relative navigation applications. We implement these libraries on the CPU, to assist in the navigation pipeline. 

The survey of Lentaris et al.\cite{lentaris2018high} evaluates the performance benchmarks in monocular and stereo vision-based solutions on various platforms. Their analysis highlight the need for hardware accelerators for high-performance avionics. Maragos et al.\cite{maragos2018evaluation} propose adaptive reconfiguration of computer vision algorithms for implementation on the European NG-MEDIUM FPGA. They evaluate feature detection and tracking algorithms for relative pose estimation with only monocular vision measurements. Camarena et al.\cite{estebanez2018fpga} explored FPGA-based multi-sensor relative navigation in space. While they evaluate monocular and stereo vision solutions on the FPGA, they adhere to onboard computer for LiDAR data pre-processing and ICP-based methods for 3D relative navigation. Also, their approach independently targets LiDAR only navigation with the execution of iterative solvers on the hardware. Apart from the geometric methods, deep learning based techniques such as PointNetLK \cite{aoki2019pointnetlk} and FMR \cite{huang2020feature} have been successfully applied to the point-cloud registration problem. However, challenges in uncooperative space missions make acquisition of large training sets difficult and expensive, making robust geometrical approaches, such as OLTAE, particularly interesting.

An FPGA accelerator for ICP-based object pose estimation through a hierarchical graph data structure is implemented by Kosuge et al.\cite{kosuge2020soc} The proposed faster ICP algorithm comes with data preprocessing overhead and may not be robust to outliers. Besides, in spacecraft navigation, if 2D image measurements are additionally available, robust solutions for pose estimation are realizable. A memory efficient FPGA accelerator for Normal Distribution Transform (NDT) method is presented by Deng et al.\cite{deng2021optimized} High-performance compute capabilities of FPGA are particularly alluring for deep learning based pose estimation solutions. Sugiura et al.\cite{sugiura2022efficient} implemented an FPGA hardware accelerator for point cloud registration using PointNetLK\cite{aoki2019pointnetlk}. They demonstrate training and testing of resource efficient hardware implementation of supervised learning methods on mid-range as well as low-cost FPGA SoCs. These investigations present a successful foray into reliable low-cost low-power hardware accelerator designs for autonomous navigation applications utilizing point cloud registration methods. 

%% file: Section3.tex


The point-cloud registration process is presented as a sequence of two steps. First, the correspondence between the point clouds is established. Next, the OLTAE algorithm is carried out to estimate the attitude and the translation vectors.


\subsection{Correspondence between two point-clouds using image features}

The determination of point correspondences between two 3D point-clouds, generated by a scanner such as a LiDAR, is a complex problem because of the numerous possible evaluations in finding a match. In this work, we assume the availability of 2D image measurements and the correspondence between the point-clouds is established with help from image features. This is advantageous because:

Finding a match between the 3D points and the 2D reference points narrows the search space significantly and addresses the slow and weak convergence issue that is otherwise persistent in an ICP like algorithm that involves matching all the points in both the point clouds. 

\begin{figure}[ht]
\centering
\includegraphics[width=0.75\textwidth]{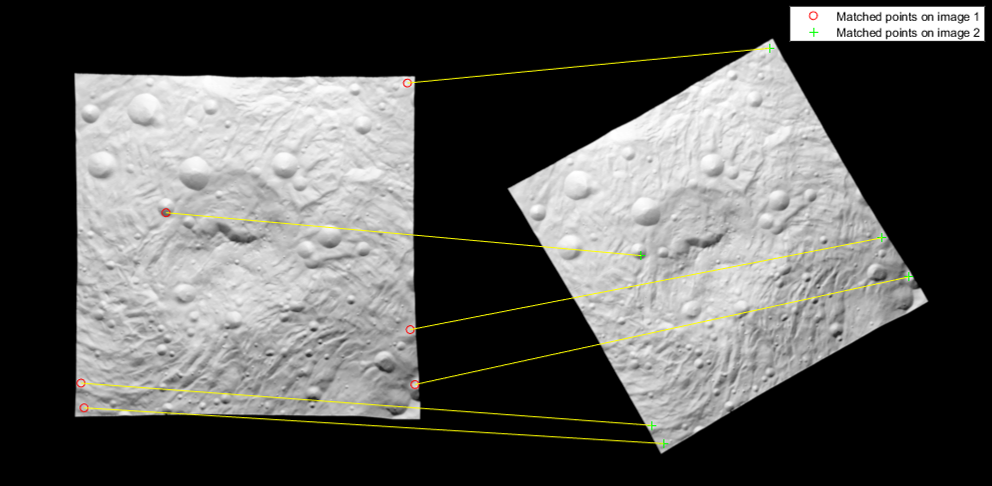}
\caption{Matched features between two image frames captured at two representative navigation events.}
\label{fig:surf_features}
\end{figure}

\begin{figure}[ht]
\centering
\includegraphics[width=0.9\textwidth]{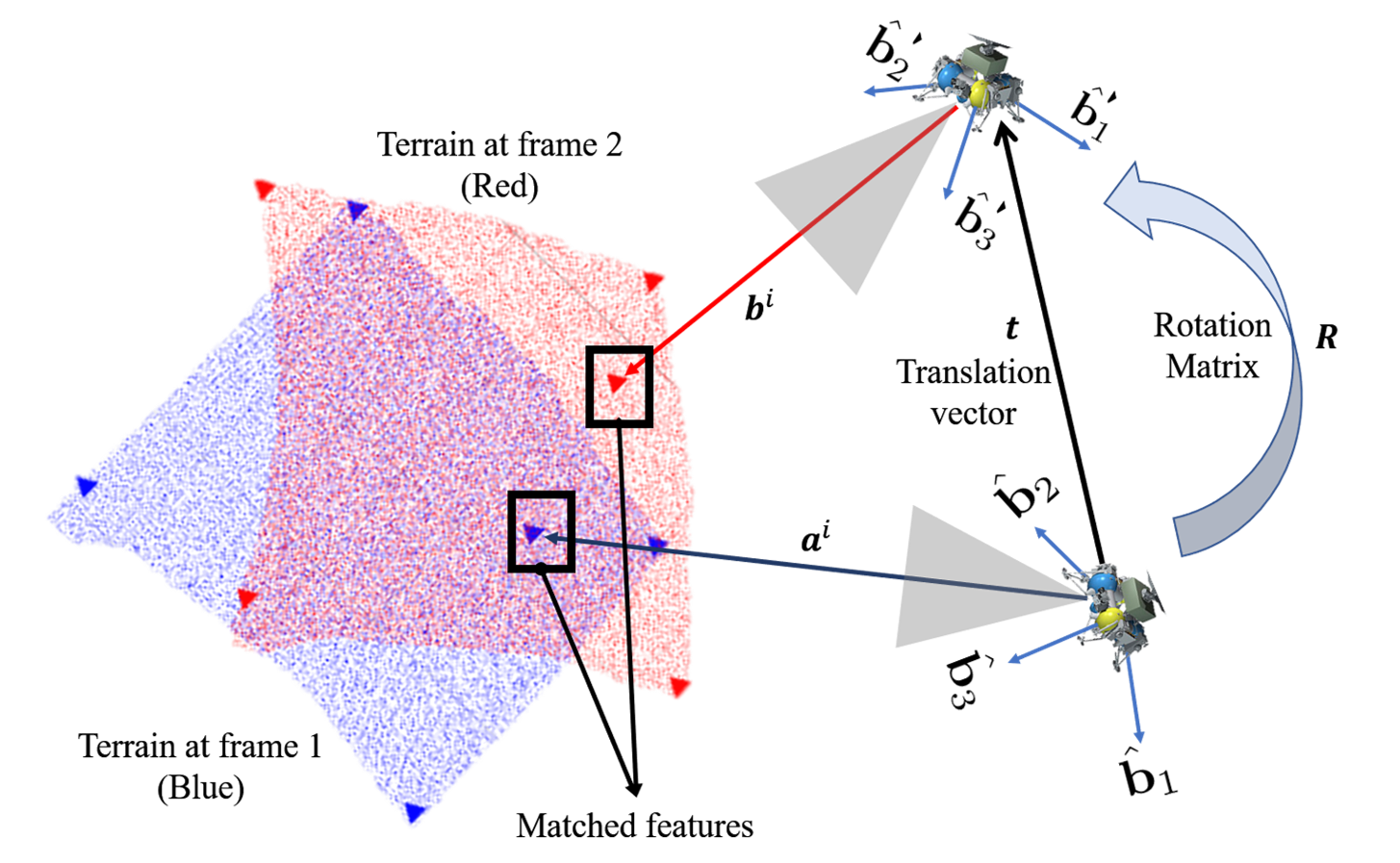}
\caption{Point-cloud of a terrain as viewed in the sensor frame of a LiDAR scanner from two different positions in the space. In this example, the features across frames 1 and 2 are matched to establish the correspondence between the two point clouds.}
\label{fig:3dMatchedFeatures_uav}
\end{figure}

Hereby, the point-cloud correspondence problem is reduced to finding corresponding points in a pair of two 2D images and then interpolate the matches to determine their respective 3D point correspondences. Correspondences between 2D images are established by detecting and matching feature points. Feature detection is based on identifying corners, edges, or other interest points that standout in the local neighborhood of the image region. For every feature that is detected, a local image patch around the feature is extracted and stored as a feature or key-point descriptor. It is by comparing these descriptors, pixel pairs of matching features are identified. A variety of feature detection algorithms are available in the literature for this task \cite{mikolajczyk2008local}. The applications involving vehicular motion, scale-invariant feature detection algorithms such as  Scale-Invariant Feature Transform (SIFT) \cite{lowe2004sift} or Speeded Up Robust Features (SURF) \cite{bay2006surf} are observed to be reliable. SURF is a blob detection program and is very robust for tasks involving object recognition and 3D reconstruction. The implementation of SURF is assisted by the use of a 64-dimensional feature descriptor for each feature detected in an image. A threshold metric for feature detection and a choice in the region of interest offered by SURF implementations assist in producing matches between two sets of images that are empirically correct. Figure \ref{fig:surf_features} shows the matched set of image features captured at two different navigation events. Figure \ref{fig:3dMatchedFeatures_uav} depicts the set of matched features re-projected in the LiDAR reference frame. We assume that the temporal association between images and LiDAR scans is pre-established to correctly match the 2D and 3D features.

The synthetic images as well as the point clouds used in this study are of Rheasilvia\footnote{CAD model of Rheasilvia: \url{https://nasa3d.arc.nasa.gov/detail/vesta-rheasilvia}}. The synthetic data is generated using a ray-tracer engine - Navigation and Rendering Pipeline for Astronautics (NaRPA) \cite{ramchander2021hardware,bhaskara_score}.

\subsection{Mathematical Formulation: Optimal Linear Translation and Attitude Estimator (OLTAE) algorithm}

Consider two point clouds mathematically related via a rigid-body transformation given by 
\begin{equation} \label{eq:oltae_transformation}
    \mathbf{b}_{i} = R  \mathbf{a}_{i} +  \mathbf{t},\; i=1,2,...n 
\end{equation}

\noindent where $\mathbf{a}_{i}$, $\mathbf{b}_{i}$ are the coordinates of the $i^{th}$ measurement point observed in two successive frames, $R$ is the $3 \times 3$ proper orthogonal rotation matrix ($R^TR = I$ and $\text{det}(R) = 1$) and, $\mathbf{t}$ is the translation vector between the successive frames. For a known correspondence between the feature coordinates ($\mathbf{a}_{i}$ and $\mathbf{b}_{i}$) in the successive frames, the point cloud registration is posed as an optimization problem that minimizes the Euclidean distance between the vector coordinates. The minimization function, $J$, is given as: 
\begin{equation} \label{eq:oltaeMinProblem}
    J = \min_{R,\mathbf{t}}  ||\mathbf{b}_{i} - R \mathbf{a}_i - \mathbf{t}||_W ^2 = \sum_{i=1}^{n} \big\{(\mathbf{b}_{i} - R \mathbf{a}_i - \mathbf{t})^T W (\mathbf{b}_{i} - R \mathbf{a}_i - \mathbf{t})\}
\end{equation}
The minimization problem in Eq. (\ref{eq:oltaeMinProblem}) could be directly solved for unknown parameters in $R$ and $\mathbf{t}$ using a weighted least-squares technique. In this work, the attitude kinematics are parameterized using the Classical Rodriguez Parameters (CRP) or Gibbs vector representation \cite{schaub1996stereographic}. CRPs facilitate posing the minimization problem via polynomial system solving, free of trigonometric functions as well as nonlinearities. Let a vector $\mathbf{q} = [q_1\; q_2\; q_3]^T$ denote the CRP, the rotation matrix $R$ in terms of $q$ can be obtained using Cayley transform as: 
\begin{equation} \label{eq:cayleyTransform}
R = (I + Q)^{-1} (I - Q)
\end{equation}
where $I \in {\rm I\!R}_{3 \times 3}$ is an identity matrix, and $Q = [\mathbf{q}\times] \in {\rm I\!R}_{3 \times 3}$ is a skew-symmetric matrix obtained from the cross-product operation on the Gibbs vector $[\mathbf{q} \times]$ 

\begin{equation}
Q = [\mathbf{q}\times] = \begin{bmatrix}
0 & -q_3 & q_2 \\
q_3 & 0 & -q_1 \\
-q_2 & q_1 & 0
\end{bmatrix}
\end{equation}  

Now Eq. (\ref{eq:oltae_transformation}) is re-written as 
\begin{gather} \label{eq:oltaeQtransformation0}
    \mathbf{b}_i = (I+Q)^{-1} (I-Q) \mathbf{a}_i + \mathbf{t}
    \\
    \label{eq:oltaeQtransformation}
    (I+Q) \mathbf{b}_i = (I-Q) \mathbf{a}_i + \mathbf{t}^*
\end{gather}

\noindent where the translation vector is redefined as $\mathbf{t}^* = (I+Q)\mathbf{t}$.

Direct manipulation of Eq. (\ref{eq:oltaeQtransformation}) simplifies it as
\begin{equation} \label{eq:oltaeRedef_trans}
     \mathbf{b}_i -  \mathbf{a}_i = -Q ( \mathbf{b}_i +  \mathbf{a}_i) + \mathbf{t}^*
\end{equation}

For ease of representation, if we define  $\mathbf{b}_i +  \mathbf{a}_i = \boldsymbol{\nu}_i$ and  $\mathbf{b}_i -  \mathbf{a}_i = \boldsymbol{\epsilon}_i$, and noting the interchangeability of the skew-symmetric operator for a matrix-vector multiplication,  
Eq. (\ref{eq:oltaeRedef_trans}) modifies into
\begin{align} \label{eq:oltaeEqlineartrans}
    \boldsymbol{\epsilon}_i &= -Q \boldsymbol{\nu}_i + \mathbf{t}^* \\
   \boldsymbol{\epsilon}_i &= [\boldsymbol{\nu} \times] \mathbf{q} + \mathbf{t}^*
\end{align}

Eq. (\ref{eq:oltaeEqlineartrans}) is a rigorously linear equation in attitude (CRP) and translation vectors, both of which can be solved directly using the linear least-squares for at least three measurement vectors. But the efficiency that the problem's structure offers is underutilized because of the higher dimensional $6 \times 6$ matrix inversions involved in this direct approach.

Restructuring of the problem in Eq. (\ref{eq:oltae_transformation}) is inspired from the work of Arun et al. \cite{arun1987least} where the centroids of the point clouds are aligned to compensate for the translation vector and only the attitude vector is solved first. The reformulation of the problem is as follows:
\begin{gather}
    \sum_{i=1}^{n} \mathbf{b}_i = R \sum_{i=1}^{n} \mathbf{a}_i + n \mathbf{t} \\  \label{eq:oltaeRestructuring}
    \frac{1}{n}\sum_{i=1}^{n} \mathbf{b}_i = R  \left(\frac{1}{n} \sum_{i=1}^{n} \mathbf{a}_i \right) + \mathbf{t}
\end{gather}

Defining $\frac{1}{n} \sum_{i=1}^{n} \mathbf{a}_i = \bar{\mathbf{a}}$ and $\frac{1}{n} \sum_{i=1}^{n} \mathbf{b}_i = \bar{\mathbf{b}}\,$, Eq. (\ref{eq:oltaeRestructuring}) gives a formula for the implicit representation of translation vector in terms of $\bar{\mathbf{a}}_i$ and $\bar{\mathbf{b}}_i$ as
\begin{equation} \label{eq:oltaeTranslation}
    \mathbf{t} = \bar{\mathbf{b}} - R \bar{\mathbf{a}}
\end{equation}
Substituting Eq. (\ref{eq:oltaeTranslation}) in Eq. (\ref{eq:oltae_transformation}) yields 
\begin{gather}
\begin{split}
    \mathbf{b}_{i} - \bar{\mathbf{b}} = R  (\mathbf{a}_{i} -  \bar{\mathbf{a}}) \\
    \delta \mathbf{b}_{i} = R  \delta \mathbf{a}_{i}
\end{split}
\end{gather}
\noindent where the definitions $\delta \mathbf{b}_{i} = \mathbf{b}_{i} - \bar{\mathbf{b}}$ and $\delta \mathbf{a}_{i} = \mathbf{a}_{i} - \bar{\mathbf{a}}$ are used. 

Considering the CRP based attitude representation $R = (I+Q)^{-1}(I-Q)$, and the formulation similar to the one described in Eqs. (\ref{eq:oltaeQtransformation0}) - (\ref{eq:oltaeEqlineartrans}), the following equations can be realized
\begin{gather}
\begin{split}
     \delta \mathbf{b}_{i} - \delta \mathbf{a}_i = -Q (\delta \mathbf{b}_{i} + \delta \mathbf{a}_i) 
     \\
     \delta \mathbf{b}_{i} - \delta \mathbf{a}_i = -[\mathbf{q} \times] \; (\delta \mathbf{b}_{i} + \delta \mathbf{a}_i) 
     \\
     \delta \mathbf{b}_{i} - \delta \mathbf{a}_i =  [(\delta \mathbf{b}_{i} + \delta \mathbf{a}_i) \times \; ] \mathbf{q}
     \end{split}
\end{gather}

By defining $ \delta \mathbf{b}_{i} - \delta \mathbf{a}_i = \delta \boldsymbol{\epsilon}_i$ and $\delta \mathbf{b}_{i} + \delta \mathbf{a}_i = \delta \boldsymbol{\nu}_i$ for the sake of simplicity, one can immediately observe a compact linear system of equation: 
\begin{gather}
    \delta \boldsymbol{\epsilon}_i = [\delta \boldsymbol{\nu}_i \times] \mathbf{q}
\end{gather}

By gathering the algebraic sum ($\delta \boldsymbol{\epsilon}$) and differences ($\delta \boldsymbol{\nu}$)  from at least three different measurement vectors, in a system matrix $H$ and a vector $\tilde{\mathbf{y}}$, we write the least-squares equation and seek the estimate for $\mathbf{q}$
\begin{gather} \label{eq:oltaeModifiedLeastSq}
     \underbrace{\begin{bmatrix}
    \delta \boldsymbol{\epsilon}_1 \\
    \delta \boldsymbol{\epsilon}_2 \\
    \vdots \\
    \delta \boldsymbol{\epsilon}_n
    \end{bmatrix}}_{\tilde{\mathbf{y}}} = \underbrace{\begin{bmatrix}
    [\delta \boldsymbol{\nu}_1 \times] \\
    [\delta \boldsymbol{\nu}_2 \times] \\
    \vdots \\
    [\delta \boldsymbol{\nu}_n \times]
    \end{bmatrix}}_{H} \mathbf{q}
    \\
    \tilde{\mathbf{y}} = H  \mathbf{q}
\end{gather}

From Eq. (\ref{eq:oltaeModifiedLeastSq}) and following the definition of the cost function in Eq. (\ref{eq:oltaeMinProblem}), the optimal estimate is obtained from the solution to the normal equation as 
\begin{gather}\label{eq:oltaeEstFinalW}
    \hat{\mathbf{q}} = (H^T W H)^{-1} H^T W \tilde{\mathbf{y}}
\end{gather}

A statistically optimal maximum likelihood estimate \cite{crassidis2011optimal} is obtained for the choice of weight matrix as the inverse of the measurement error covariance, $\Sigma$ which in turn is a consequence of uncertainty in the image feature extraction process. Thus, the least-squares problem transforms to 
\begin{equation} \label{eq:oltaeEstFinal0}
     \hat{\mathbf{q}} = (H^T \Sigma^{-1} H)^{-1} H^T \Sigma^{-1} \tilde{\mathbf{y}}
\end{equation}

\subsection{Modified closed-form solution to the OLTAE algorithm}

While Eq. (\ref{eq:oltaeEstFinal0}) is the least-squares solution to the attitude estimation problem, the structure and make of the system matrix $H$ offers a scope of reducing the dimensionality of matrix operations involved. A glance at the system matrix $H$ in Eq. (\ref{eq:oltaeModifiedLeastSq}) indicates its composition with skew-symmetric sub-matrices $ [\delta \boldsymbol{\nu}_i \times]$. Using the properties of skew-symmetric matrices, the terms in the Eq. (\ref{eq:oltaeEstFinal0}) are re-written as explained below. First, define the skew-symmetric matrix in an equivalent yet compact form as
\begin{equation} \label{eq:oltaeskewSym}
    [\delta \boldsymbol{\nu}_i \times] = [\tilde{\mathbf{s}}_i]
\end{equation}
\noindent and define the measurement covariance matrix $\Sigma$ as 
\begin{equation} \label{eq:oltaecovariance}
    \Sigma = \begin{bmatrix}
    \sigma_1^2 & & \\
    & \ddots & \\
    & & \sigma_n^2
    \end{bmatrix}
\end{equation}
\noindent Using Eqs. (\ref{eq:oltaeskewSym}) and (\ref{eq:oltaecovariance}), the first part of Eq. (\ref{eq:oltaeEstFinal0}) is expanded as  
\begin{align}
    H^T \Sigma^{-1} H &= \begin{bmatrix}
       -[\tilde{\mathbf{s}}_1] & -[\tilde{\mathbf{s}}_2] & \hdots & -[\tilde{\mathbf{s}}_n]
    \end{bmatrix}  \begin{bmatrix}
    \frac{1}{\sigma_1^2} & & \\
    & \ddots & \\
    & & \frac{1}{\sigma_n^2}
    \end{bmatrix} \begin{bmatrix}
       [\tilde{\mathbf{s}}_1] \\ [\tilde{\mathbf{s}}_2] \\ \vdots \\ [\tilde{\mathbf{s}}_n]
    \end{bmatrix} \notag
    \\
    &= - \sum_{j = 1}^{n} \; [\tilde{\mathbf{s}}_j] \; \frac{1}{\sigma_j^2} I_{3 \times 3} \; [\tilde{\mathbf{s}}_j] \notag
    \\
    &= - \sum_{j = 1}^{n} \; \frac{1}{\sigma_j^2} \; [\tilde{\mathbf{s}}_j]^2 \notag
    \\ 
    \label{eq:oltae_outerInner}
    &= \sum_{j = 1}^{n} \; \frac{1}{\sigma_j^2} \; [\mathbf{s}_j^T \mathbf{s}_j I_{3 \times 3} - \mathbf{s}_j \mathbf{s}_j^T]
\end{align}
\noindent where the above equation embeds the squared skew-symmetric matrix identity written in its elegant form involving only vector products as: 
\begin{equation} \label{eq:oltae_vectorArithmetic}
    [\tilde{\mathbf{s}}_j]^2 = -(\mathbf{s}_j^T \mathbf{s}_j I_{3 \times 3} - \mathbf{s}_j \mathbf{s}_j^T)
\end{equation}
\noindent The second part of Eq. (\ref{eq:oltaeEstFinal0}) is the matrix vector product, written as
\begin{align}
   H^T \Sigma^{-1} \tilde{\mathbf{y}} &= \begin{bmatrix}
       -[\tilde{\mathbf{s}}_1] & -[\tilde{\mathbf{s}}_2] & \hdots & -[\tilde{\mathbf{s}}_n]
    \end{bmatrix}  \begin{bmatrix}
    \frac{1}{\sigma_1^2} & & \\
    & \ddots & \\
    & & \frac{1}{\sigma_n^2}
    \end{bmatrix} \begin{bmatrix}
       \mathbf{\tilde{y}}_1 \\ \mathbf{\tilde{y}}_2 \\ \vdots \\ \mathbf{\tilde{y}}_n
    \end{bmatrix} \notag
    \\
    &= - \sum_{j = 1}^{n} \; [\tilde{\mathbf{s}}_j] \; \frac{1}{\sigma_j^2} \; \mathbf{\tilde{y}}_j \notag
    \\
    &= - \sum_{j = 1}^{n} \; \frac{1}{\sigma_j^2} \; [\tilde{\mathbf{s}}_j] \;\mathbf{\tilde{y}}_j \notag
    \\ \label{eq:oltae_crossProduct}
    &=  - \sum_{j = 1}^{n} \; \frac{1}{\sigma_j^2} \; [\mathbf{s}_j \times \mathbf{\tilde{y}}_j] 
\end{align}
\noindent where the product between a skew-symmetric matrix ($[{\mathbf{u}} \times]$) and a vector ($\mathbf{v}$) is identified as a cross product between two vectors ($\mathbf{u}$ and $\mathbf{v}$) such that $[{\mathbf{u}} \times] \mathbf{v} = {\mathbf{u}} \times \mathbf{v}$.

The modifications to the OLTAE algorithm in Eq. (\ref{eq:oltaeEstFinal0}) with centroid alignment transforms the complete array of involved matrix manipulations into a lower-dimensional $3 \times 3$ operations which are computationally very efficient given the cubic increment in the cost ($\mathcal{O}(n^3)$) of matrix multiplications and inversions. Also, from a hardware implementation point-of-view, working with the lower dimensional matrix operations is extremely resource-efficient.

Ultimately, the equation for the least-squares algorithm used in this work, is given below: 
\begin{equation}\label{eq:oltaeEstFinal}
     \hat{\mathbf{q}} = -\left(  \sum_{j = 1}^{n} \; \frac{1}{\sigma_j^2} \; [\mathbf{s}_j^T \mathbf{s}_j I_{3 \times 3} - \mathbf{s}_j \mathbf{s}_j^T] \right) ^{-1} \left( \sum_{j = 1}^{n} \; \frac{1}{\sigma_j^2} \; [\mathbf{s}_j \times \mathbf{\tilde{y}}_j] \right) 
\end{equation}

From the estimate of the Gibbs vector $\hat{\mathbf{q}}$, the attitude matrix is computed using the Cayley Transform in Eq. (\ref{eq:cayleyTransform}). Finally, the translation vector is retrieved from the definition in Eq. (\ref{eq:oltaeTranslation}) to complete the 6-DOF pose estimation process. Matrix operations wise, the algorithm recursively computes addition on $2n$ number of $3 \times 3$ matrices obtained in Eqs. (\ref{eq:oltae_outerInner}) and (\ref{eq:oltae_crossProduct}), a matrix inversion and a matrix multiplication as indicated in Eq. (\ref{eq:oltaeEstFinal}). To obtain the matrices, the design involves vector outer product, inner product, and cross product computations as succinctly expressed in Eq. (\ref{eq:oltae_crossProduct}). The modified OLTAE solution solves scalability by handling numerous feature correspondences without hardware modifications, and is suitable for low-latency pose estimation due to its lower dimensional operations.

\subsection{Software simulation pipeline}
\comment{
\begin{table}[ht]
\setlength{\arrayrulewidth}{0.5mm}
\setlength{\tabcolsep}{14pt}
\renewcommand{\arraystretch}{1.25}
\centering
 \begin{tabular}{c c} 
 \hline
 Operation & Complexity \\ [0.5ex] 
 \hline
Inner Product &  $\mathcal{O}(n)$ \\
Cross Product &  $\mathcal{O}(n)$ \\
Outer Product &  $\mathcal{O}(n^2)$ \\
Matrix Multiplication &  $\mathcal{O}(n^3)$ \\
Matrix Inversion  &  $\mathcal{O}(n^3)$ \\
 \hline
 \end{tabular}
  \caption{\label{tab:oltae_timeComplexity} Time complexities of operations involved in the OLTAE algorithm.}
\end{table}
This high-performance variant of the OLTAE algorithm is exploited in this work to realize the hardware implementation of it. In the remainder of this article, the design and implementation details of the hardware architecture for the OLTAE algorithm is presented. 
}

\begin{figure}[ht]
\centering
\includegraphics[width=0.6\textwidth]{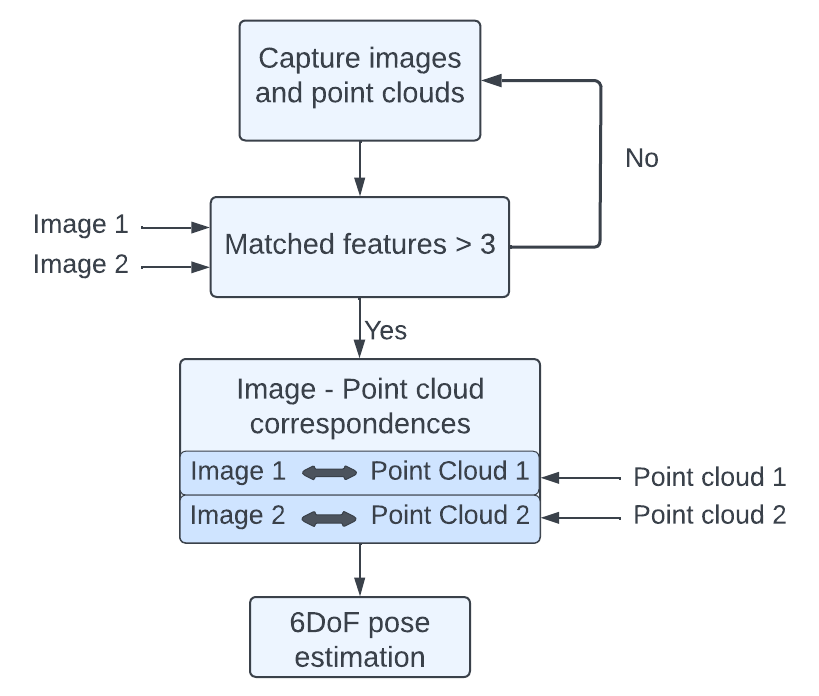}
\caption{Software simulation pipeline for the relative pose estimation using OLTAE.}
\label{fig:flowChartSimPipeline}
\end{figure}

In this vision-aided point cloud registration process, first, the correspondence between 2D image features and the 3D point cloud is established. The images are captured by a high frame rate camera sensor or by transforming the 3D point cloud to a 2D bearing angle image \cite{lin2017novel}. These correspondences are then tracked across frames to estimate pose transformation between the 3D point clouds. The flow chart of the simulation pipeline is shown in Figure \ref{fig:flowChartSimPipeline}. 

In the first step, both the images and point cloud data are registered in their respective sensor frames. If required, these are transformed to a common body-frame subject to respective sensor placements. Next, the features common to the image sets in the representative captures are detected and extracted using the image feature extraction and matching tools. Then, the 3D location of the detected features is estimated from the point cloud and the matched image feature data. Prior to identifying the 3D location of the feature from its corresponding image pixel coordinates, availability of at least three uniquely matched features is to be ensured to avoid an otherwise singularity problem. Also, because a matched set of feature coordinates are usually floating-point numbers, the use of interpolation techniques \cite{ramchander2021hardware} is essential to search for the 3D location in the integer-type pixel coordinates. 

\comment{
\begin{figure}[ht]
\centering
\subfloat[Image at camera location 1.]{\label{fig:o_imSURF1}\includegraphics[width=0.4\textwidth]{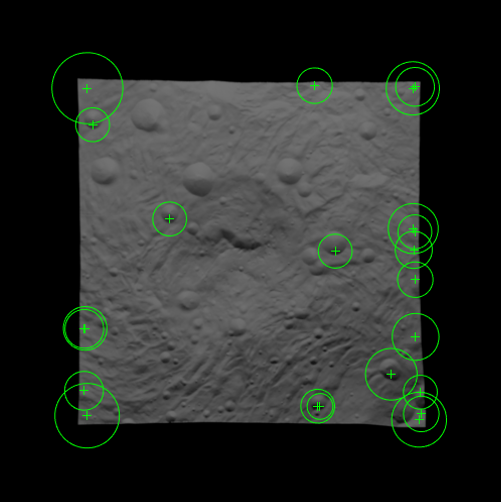}}\qquad
\subfloat[Image at camera location 2.]{\label{fig:o_imSURF2}\includegraphics[width=0.4\textwidth]{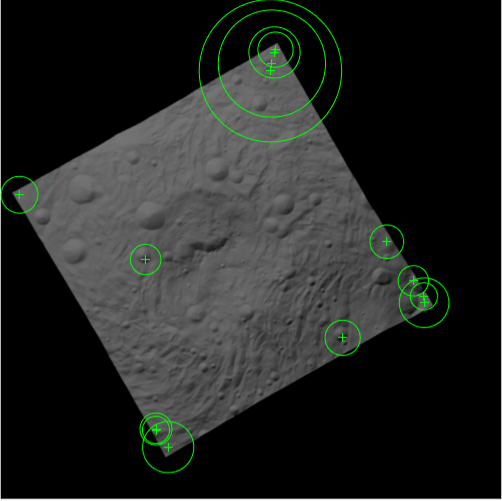}}\\
\subfloat[Matched features between images captured at two different camera locations.]{\label{fig:o_surfMatched}\includegraphics[width=1\textwidth]{figures/o_surfMatchedFeatures1.png}}%
\caption{SURF feature detection and extraction, shown on two images of a terrain named Rheasilvia. Image 1 is taken from a reference position and image 2 is taken after the camera is transformed to a different position. The matched features between images 1 and 2 are also shown.}
\label{fig:surfImageMatch}
\end{figure}
}


%% file: section4.tex



\begin{figure}[ht]
\centering
\includegraphics[width=0.6\textwidth]{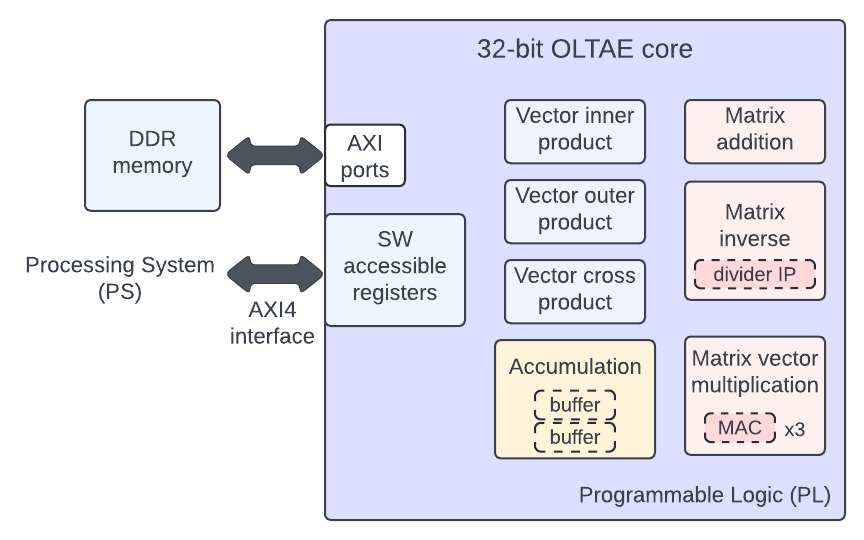}
\caption{OLTAE HW/SW co-design: The modules on the OLTAE core on the PL evaluate the least squares based attitude estimation logic (Eq. \ref{eq:oltaeEstFinal}). The PS facilitates the communication of measurement data and estimation results via Advanced eXtensible Interface (AXI4) bus interface.}
\label{fig:oltae_core}
\end{figure}
\FloatBarrier
 
\begin{figure}[ht]
\centering
\includegraphics[width=1\textwidth]{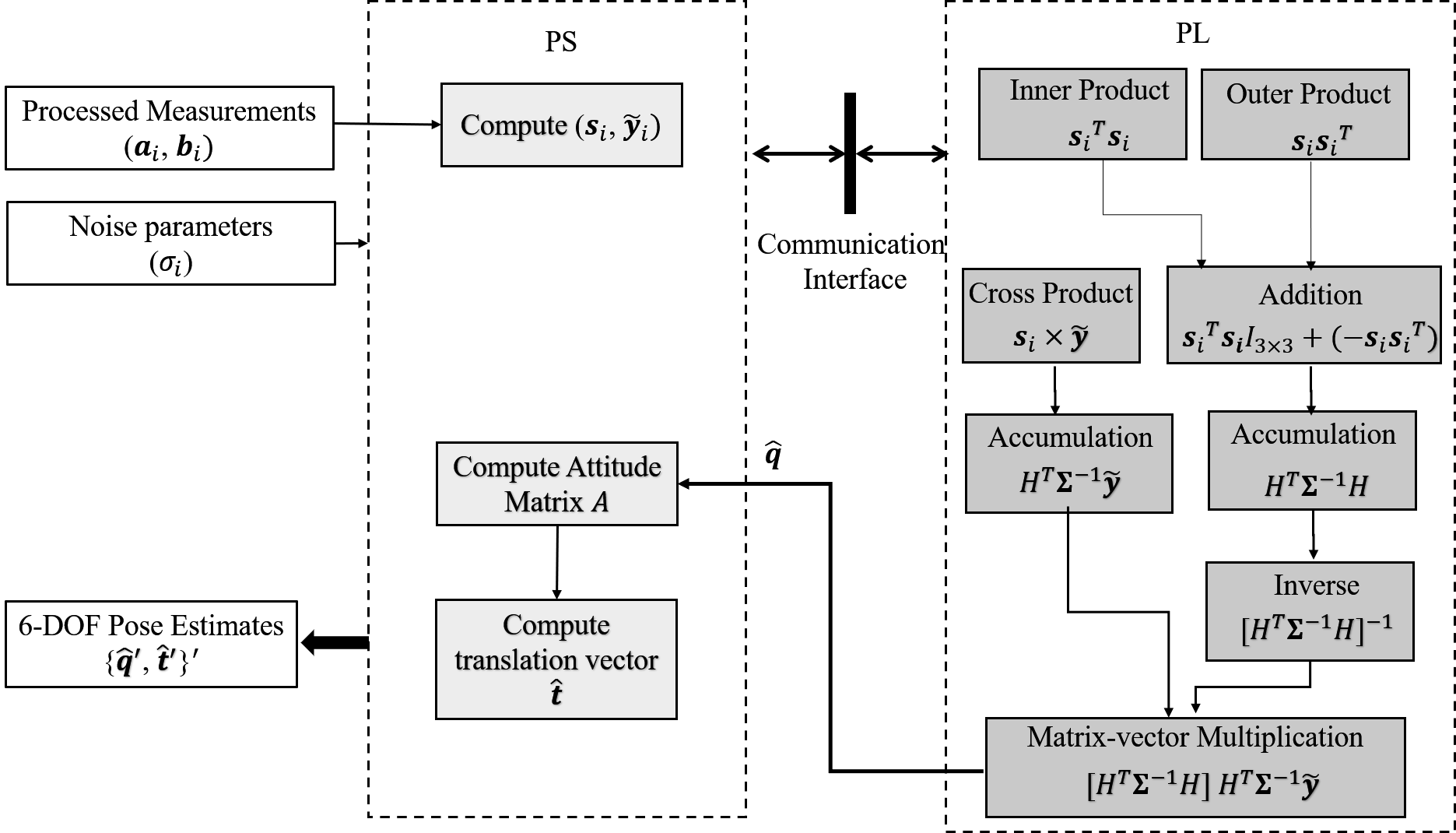}
\caption{This figure illustrates the hardware/software co-design and the flow of operations for the 6-DOF pose estimation using OLTAE algorithm.}
\label{fig:oltae_coreFlow}
\end{figure}

The architecture of the OLTAE is a hardware/software (HW/SW) co-design that evaluates the least squares based pose estimation on the PL, while the point cloud registration and data processing are handled by the PS, as shown in Figure \ref{fig:oltae_core}. The PS and the OLTAE core on the PL implement the sequence of linear algebra operations for pose estimation as shown in Figure \ref{fig:oltae_coreFlow}. The OLTAE core executes all the matrix operations required for a 3-DOF attitude vector (CRP) computation from a set of processed measurements that are expected from the software. The core relies on the processing system (PS) to obtain the simulated point-cloud correspondences. These are fed to the programmable logic (PL) as processed measurements. A natural choice of design would be to exploit the parallel computation of the FPGA for the intensive feature extraction methods as well. But this work is targeted towards OLTAE implementation for acceleration of the registration process, independent of the techniques that furnish the processed measurement data. 

The hardware modules evaluate matrix addition, inversion, vector inner, outer, and cross products, and matrix-vector multiplications. A 2D topology of multiplier and accumulator (MAC) units, known as Systolic array architecture (SAA) \cite{kung1982systolic}, is implemented to evaluate a pipelined architecture for matrix multiplication. Matrix-vector multiplication is computed using MAC units and the same operating principle as the SAA. Matrix inversion is evaluated using the Cramer's rule and Xilinx's divider IP core \cite{dividerIPXilinx}. 

Additionally, the incoming stream of data is accumulated on the PL for appropriately capturing the measurements for the least squares estimation. The PS controls the data flow, to and from the PL, using software-accessible registers and Advanced eXtensible Interface (AXI4). The PL reads the data from these registers and writes the operational status and estimation results. 

\subsection{State machine}\label{subsec:stateMachine}
\begin{figure}[ht]
\centering
\includegraphics[width=0.54\textwidth]{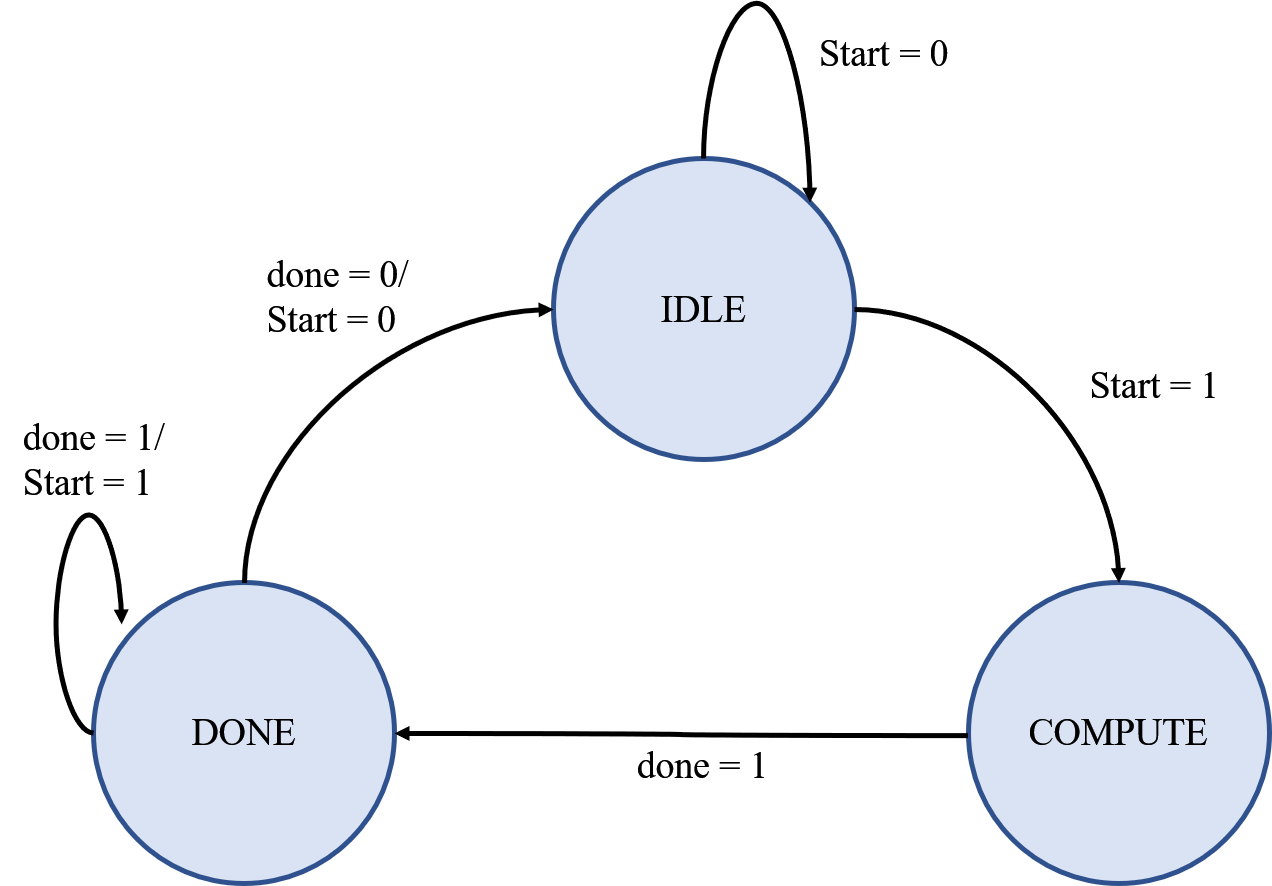}
\caption{State Machine diagram for the operation of OLTAE core.}
\label{fig:stateMachine3}
\end{figure}
Data flow control between the PS and the PL is achieved via a state machine shown in Figure \ref{fig:stateMachine3}. The operation of the state machine is as follows: the PL is set in an \textit{IDLE} state until a \textit{start} is asserted by the PS when it begins to send data over the software accessible registers. The PL switches to a \textit{COMPUTE} state when it implements the OLTAE logic. Upon completion of the attitude estimation, the PL writes the result to the memory and proceeds to a \textit{DONE} state. The PS resets the appropriate signals on the PL for the next cycle of OLTAE operations.


%% file: section5.tex


The OLTAE core is implemented on the programmable logic of the FPGA fabric, with the processing system providing the simulated inputs that correspond to the 2D keypoint features. MATLAB's 64-bit double-precision floating-point execution of the OLTAE algorithm is taken as a golden reference for evaluating the performance of the 32-bit fixed-point FPGA implementation of the OLTAE core. To prevent finite word length effects, the incoming data is appropriately scaled to fit and to accurately represent the input in the 32 bits of precision available (1 sign bit, 15 integer bits, 16 fractional bits). To this extent, uniformly scaling the input vectors is found to be adequate to faithfully represent the numbers in the limited precision range. The scaling adopted for the processed inputs is as follows: 

\noindent The processed measurement vector $\mathbf{s}_j$ is scaled by a constant factor $\alpha$ and the vector $\mathbf{y}_j$ by a constant $\beta$. With the application of scaling terms, the attitude estimation vector $\hat{\mathbf{q}}$ in Eq. (\ref{eq:oltaeEstFinal}) manipulates into:
\begin{gather}
    {\mathbf{s}_j^{'}} = \alpha \mathbf{s}_j
    \\
    \mathbf{y}_j^{'} = \beta \mathbf{y}_j
        \\
    \hat{\mathbf{q}}^{'} = -\left(  \sum_{j = 1}^{n} \; \frac{1}{\sigma_j^2} \; [{\mathbf{s}_j^{'}}^T \mathbf{s}_j^{'} I_{3 \times 3} - \mathbf{s}_j^{'} {\mathbf{s}_j^{'}}^T] \right) ^{-1} \left( \sum_{j = 1}^{n} \; \frac{1}{\sigma_j^2} \; [\mathbf{s}_j^{'} \times {\tilde{\mathbf{y}_j^{'}}}] \right) 
\end{gather}
\noindent from which, upon simplification, we get
\begin{align}
      \hat{\mathbf{q}}^{'} &= - \frac{\beta}{\alpha} \left(  \sum_{j = 1}^{n} \; \frac{1}{\sigma_j^2} \; [\mathbf{s}_j^T \mathbf{s}_j I_{3 \times 3} - \mathbf{s}_j \mathbf{s}_j^T] \right) ^{-1} \left( \sum_{j = 1}^{n} \; \frac{1}{\sigma_j^2} \; [\mathbf{s}_j \times \mathbf{\tilde{y}}_j] \right) 
      \\
      \hat{\mathbf{q}}^{'} &= \frac{\beta}{\alpha} \hat{\mathbf{q}}
\end{align}
\noindent ultimately, the manipulated estimation vector is retrieved as 
\begin{equation}
    \hat{\mathbf{q}} = \frac{\alpha}{\beta} \hat{\mathbf{q}}^{'} 
\end{equation}

\subsection{Simulation and functional verification}

A spacecraft landing maneuver in the terrain relative navigation (TRN) paradigm is chosen for the validation and performance benchmarking of the OLTAE algorithm. Here, the spacecraft is configured to approach the landing site in a dual-axis translation and a single-axis rotation trajectory\footnote{Trajectory design: \url{https://drive.google.com/drive/folders/1jUjAFNykEv7xDQnlIIItOT3751njlkeB}}. The maneuver follows a simplistic approach trajectory with constant linear and angular velocities. Further, the sensor positioning is such that, a virtual LiDAR scanner is configured on the spacecraft so that it could scan the landing terrain in its field of view and return the point cloud data as its output. A virtual monocular camera system, having the same field of view and resolution as that of the LiDAR, is attached alongside the scanner to return the 2D image data as its output. The entire maneuver is carried out on a set of 25 frames from the LiDAR and the camera, each generating output at a rate of 1 frame per second. In a real-time application, a high frame rate camera would produce imagery for fusion with LiDAR measurements. Alternately, 2D bearing angle images may be produced from the 3D point clouds.

\begin{figure}[ht]
\centering
\includegraphics[width=1\textwidth]{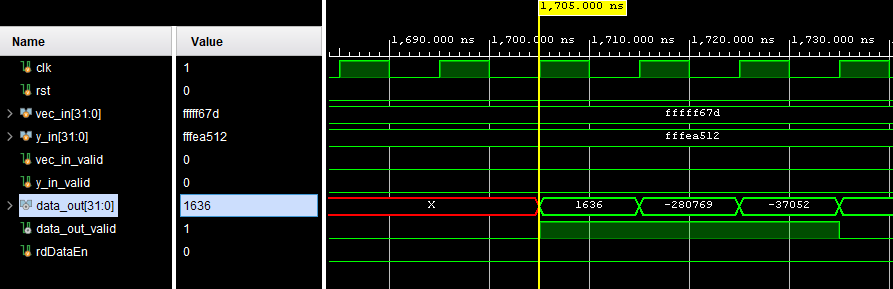}
\caption{Logic waveform of the OLTAE core simulation.}
\label{fig:oltae_simSnapshot}
\end{figure}

The OLTAE algorithm is simulated on the Vivado simulator environment for pre-silicon functional verification. The logic waveform of the OLTAE core simulation is shown in Figure \ref{fig:oltae_simSnapshot}. The simulated sensor dataset configured for processing on the PL was loaded into a testbench and the OLTAE core is simulated with the synthetic data to mimic data received from actual sensors. The estimated states (CRPs) from the data generated from a model navigation scenario are read into MATLAB for analysis. The simulation is carried out for a sequence of 25 frames in a terrain relative navigation (TRN) setting.  

\subsection{Performance analysis}

The OLTAE core is structured in a hierarchical design, with the primitive modules instantiated in a top module. In other words, the top module is defined to invoke the primitives to compute the required matrix-vector manipulations according to the design flow in Figure \ref{fig:oltae_coreFlow}. Each primitive module in the RTL design of the OLTAE core is simulation tested for logical validity before it is integrated into the top module. Post the RTL verification, the design is synthesized and implemented for simulating the system and measurement data on the Zynq Z-7020 FPGA SoC platform. The results of the simulation are observed on a logic waveform for comparison with the golden reference, i.e., results from MATLAB. 
\begin{figure}[hp]
\centering
\includegraphics[width=0.6\textwidth]{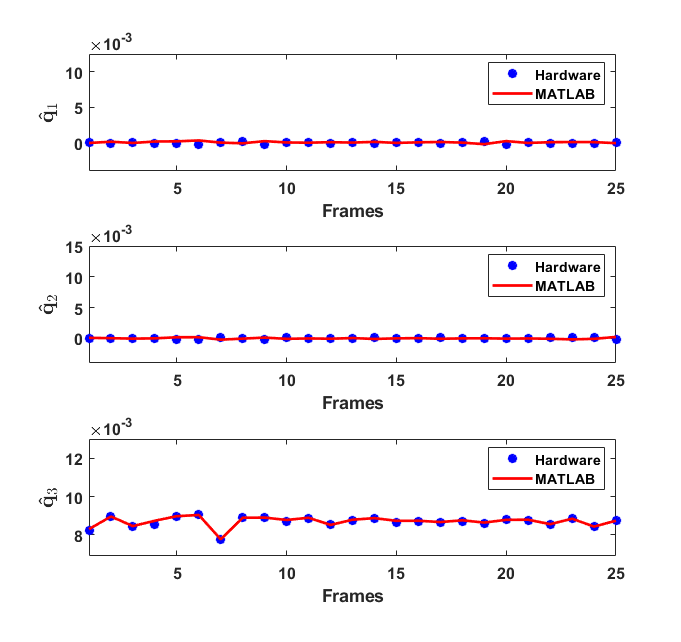}
\caption{Estimates of attitude parameters (CRP) of the OLTAE core in comparison with the MATLAB results.}
\label{fig:oltae_HWSW_q}
\end{figure}

\begin{figure}[hp]
\centering
\includegraphics[width=0.6\textwidth]{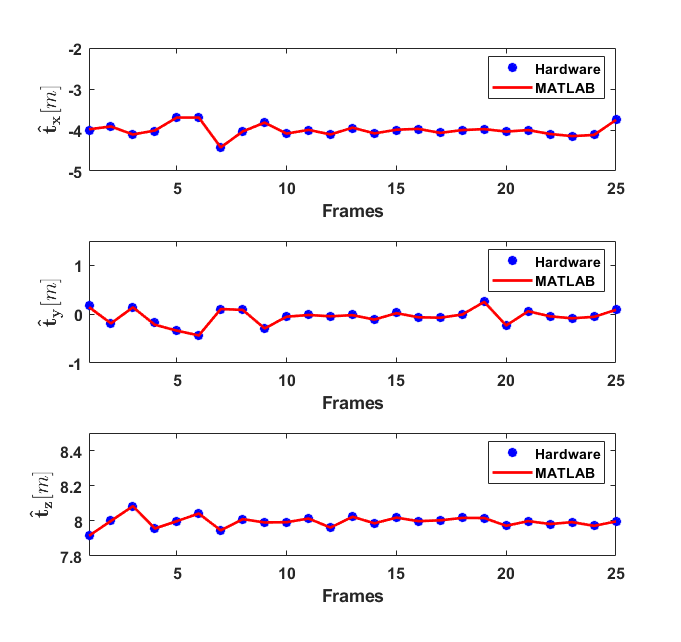}
\caption{Estimates of translation vector from the OLTAE core in comparison with the estimates from MATLAB implementation of the OLTAE algorithm.}
\label{fig:oltae_HWSW_t}
\end{figure}
%
The simulation results that the core produced (within working precision), for all the 25 frames of the TRN exercise, are compared against the results from software implementation (MATLAB). As illustrated in Figure \ref{fig:oltae_HWSW_q}, the estimates of the attitude parameters from the hardware simulation are overlaid on top of the CRP estimates from the MATLAB. The comparison between the estimates of the translation vector obtained from the OLTAE core against the software implementation, is shown in Figure \ref{fig:oltae_HWSW_t}. The plots indicate a close agreement between the hardware and software simulations.
\begin{figure}[hp]
\centering
\includegraphics[width=0.6\textwidth]{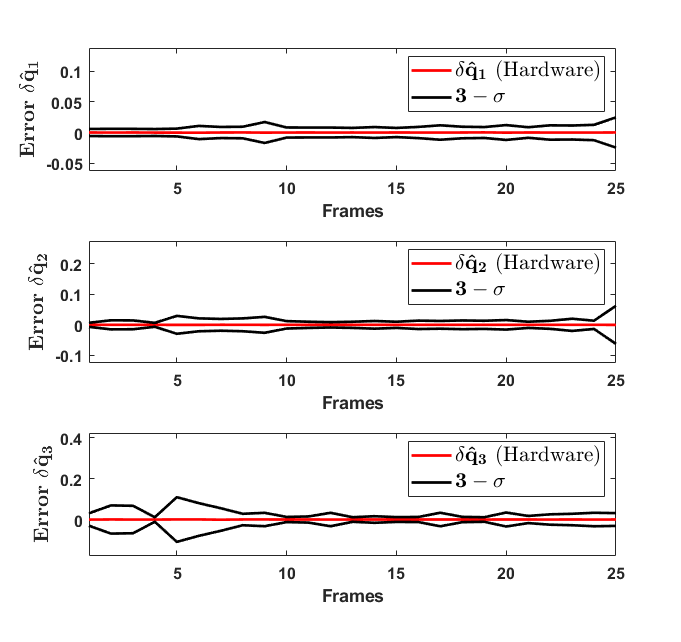}
\caption{Errors along three axes for the attitude estimates obtained from the OLTAE core (Hardware).}
\label{fig:oltae_HWSW_qerror}
\end{figure}

\begin{figure}[hp]
\centering
\includegraphics[width=0.6\textwidth]{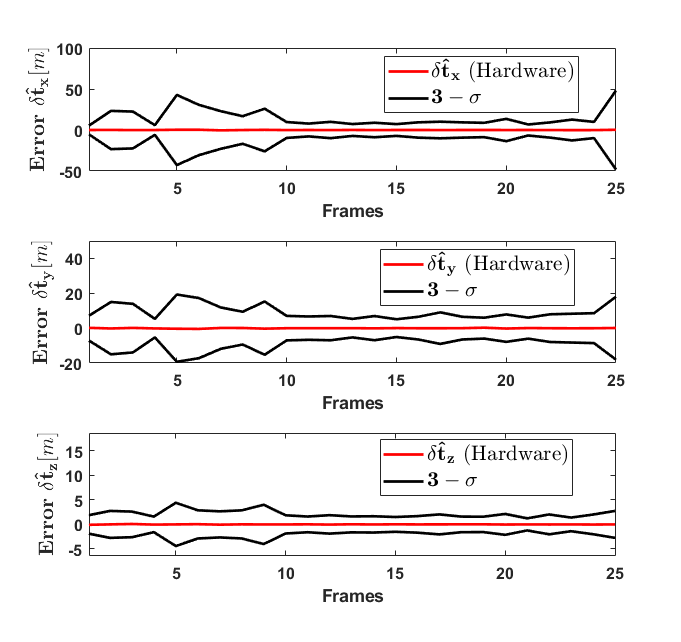}
\caption{Errors along three axes for the translation vector estimates recovered from the OLTAE core (Hardware).}
\label{fig:oltae_HWSW_terror}
\end{figure}

Linear covariance analysis for the errors obtained from the estimation process shows that the errors are statistically bounded within their respective $3-\sigma$ limits. The errors in the attitude vector are shown in Figure \ref{fig:oltae_HWSW_qerror} while the errors in the translation vector are shown in Figure \ref{fig:oltae_HWSW_terror}.

\begin{figure}[h]
\centering
\includegraphics[width=0.6\textwidth]{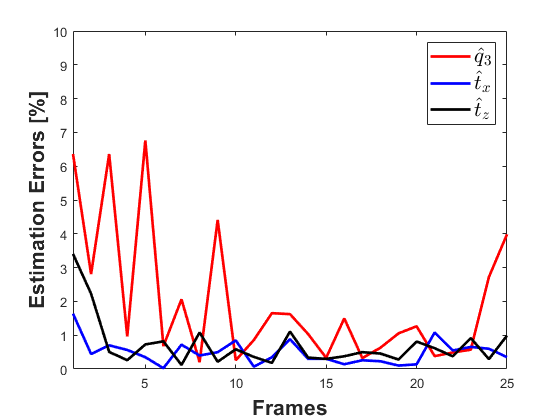}
\caption{Percentage of relative state errors between the FPGA and the MATLAB estimates. The errors captured along the states [$q_3,\, t_x,\, t_z$] (where motion is induced).}
\label{fig:olae_matlabFPGA}
\end{figure}

Relative error percentages between the FPGA and the MATLAB implementations of the OLTAE logic, at every frame, are captured in Figure \ref{fig:olae_matlabFPGA}. The state estimates are observed to have a maximum numerical deviation of $7\%$. The error statistics are applicable to the specified application data with an arbitrary scaling used to represent fixed-point integers on the hardware. Optimal scaling strategies to further reduce this residual are to be explored. The study of error and covariance plots indicates that the OLTAE core can faithfully replicate the software-based implementation of the OLTAE algorithm within the working precision. The attitude estimation takes only $1.7$ micro seconds on the Zynq FPGA-SoC with $100$ MHz clock, while the same process consumes $2.1$ milliseconds on Intel i7-7800X CPU clocked at $3.5$ GHz. 

Table \ref{tab:oltae_resources} shows the utilization of resources of the implemented OLTAE core on the FPGA programmable logic. The high throughput matrix operations in the core implement digital signal processor (DSP) blocks for high-speed parallel multiplications. This design choice may be relaxed to optimize on the area and save resources for additional logic. 

\begin{table}[htbp]
\caption{Implementation requirements for the OLTAE core.}
\begin{center}
\begin{tabular}{|c|c|c|c|}
\hline
{Resource} & {Available} & {Utilization} & {Utilization \%} \\
\hline
Lookup tables (LUT) & 53200 & 5545 & $10.42$ \\
\hline
LUTRAM & 17400 & 153 & $0.88$ \\
\hline
Flip-flops (FF) & 106400 & 7597 & $7.14$ \\
\hline
DSP & 220 & 137 & $62.27$ \\
\hline
\end{tabular}
\label{tab:oltae_resources}
\end{center}
\end{table}

In conclusion, the comparison results demonstrate that the design for the OLTAE core offer reliability akin to that of the software based implementation while providing performance acceleration to the state estimation process.


%% file: section6.tex

An FPGA acceleration framework for relative pose estimation using multi-sensor (camera, LiDAR) measurements and a lightweight estimation algorithm is developed in this paper. By using Classical Rodrigues Parameters (CRPs) for attitude representation, the 6-DoF pose is estimated as a solution to a linear system of equations (linear least squares). Centroid alignment of point cloud measurements makes the point registration – a 3-DoF attitude estimation problem. By utilizing the attitude estimates, the 3-DoF translation vector is recovered next. Using vector algebra, the linear system of equations is further condensed into a lower-dimensional, implementation-friendly state estimation problem. This process is proposed as an Optimal Linear Translation and Attitude Estimator (OLTAE) algorithm. 

A high-performance HW/SW co-design is proposed to accelerate the registration process. The co-design is shown to achieve the performance and accuracy requirements for real-time navigation with a latency of $1.7 \mu$s per cycle of attitude estimation (with preprocessed input data). Simulated vision measurements from a representative trajectory are utilized to verify the functioning of the OLTAE logic on the FPGA. The state estimation on the FPGA, for an arbitrarily chosen data-scaling procedure, yielded a maximum deviation of about $7\%$ from a MATLAB implementation of the same method. The performance benefits of the FPGA acceleration indicates the utility of customized navigation filters as well as edge computing for real-time navigation and guidance.  

The FPGA framework is designed as a standalone IP core for interfacing with external modules for image data processing. This development is limited to the evaluation of estimation process on the FPGA programmable logic, while we rely on the processing system for feature detection and matching. Future efforts may be directed at embedding computationally expensive image processing pipelines, with outlier rejection schemes, for the execution of complete, custom navigation solutions. 